%% file: main.tex
\documentclass[conference]{IEEEtran}
\IEEEoverridecommandlockouts
\usepackage{cite}
\usepackage{amsmath,amssymb,amsfonts}
\usepackage{algorithmic}
\usepackage{graphicx}
\usepackage{textcomp}
\usepackage{xcolor}
\usepackage{bm}
\usepackage[ruled,linesnumbered,algo2e]{algorithm2e}
\usepackage{caption}
\def\BibTeX{{\rm B\kern-.05em{\sc i\kern-.025em b}\kern-.08em
    T\kern-.1667em\lower.7ex\hbox{E}\kern-.125emX}}

\begin{document}

\newcommand{\SBELcomment}[1]{\textbf{\textcolor{red}{{#1}}}}

\title{Instance Performance Difference: A Metric to Measure the Sim-To-Real Gap in Camera Simulation}

\author{
	\IEEEauthorblockN{Bo-Hsun Chen}
	\IEEEauthorblockA{
		\textit{Simulation-Based Engineering Lab} \\
		\textit{University of Wisconsin -- Madison}\\
		Madison, WI, USA
	}
	\and
	\IEEEauthorblockN{Dan Negrut}
	\IEEEauthorblockA{
		\textit{Simulation-Based Engineering Lab} \\
		\textit{University of Wisconsin -- Madison} \\
		Madison, WI, USA
	}
}

\maketitle

\begin{abstract}
In this contribution, we introduce the concept of \textit{Instance Performance Difference (IPD)}, a metric designed to measure the gap in performance that a robotics perception task experiences when working with real vs. synthetic pictures. By pairing synthetic and real instances in the pictures and evaluating their performance similarity using perception algorithms, IPD provides a targeted metric that closely aligns with the needs of real-world applications. We explain and demonstrate this metric through a rock detection task in lunar terrain images, highlighting the IPD's effectiveness in identifying the most realistic image synthesis method. The metric is thus instrumental in creating synthetic image datasets that perform in perception tasks like real-world photo counterparts. In turn, this supports robust sim-to-real transfer for perception algorithms in real-world robotics applications.
\end{abstract}


\section{Introduction}
\label{sec:intro}
\input{sections/intro.tex}

\section{Related Work}
\label{sec:related_works}
\input{sections/related_works.tex}

\section{Method}
\label{sec:method}
\input{sections/method.tex}

\section{A Use Case}
\label{sec:use_case}
\input{sections/use_case.tex}

\section{Conclusions}
\label{sec:conclusions}
\input{sections/conclusions.tex}

\bibliographystyle{IEEEtran}
\bibliography{BibFiles/refsSensors,BibFiles/refsML-AI,BibFiles/refsAutonomousVehicles,BibFiles/refsChronoSpecific,BibFiles/refsSBELspecific,BibFiles/refsMBS,BibFiles/refsCompSci,BibFiles/refsTerramech,BibFiles/refsFSI,BibFiles/refsRobotics,BibFiles/refsDEM,BibFiles/refsGraphics}

\end{document}

%% file: sections/intro.tex
There exist several commonly-used metrics used to evaluate the quality of synthetic images and define an associated simulation-to-reality (sim-to-real) gap that pertains to how humans respond to these images \cite{wang2004image, salimans2016improved, heusel2017gans, zhang2018unreasonable, binkowski2018demystifying}. These metrics are not suitable for the case when the synthetic images, which have individual real counterparts, are used for training perception neural networks (NNs) employed in robotic systems. Specifically in such a case, each synthetic image can be one-to-one paired with a corresponding real photo, but pixel positions of each object are not aligned perfectly between the real and synthetic pictures due to slight variations in the camera and/or object poses. Note that because synthetic images are used to train and test NNs for visual perception tasks, we care more about algorithmic performance than visual fidelity in a human perception sense. 

From the perspective of simulation, we propose an \textit{Instance Performance Difference (IPD)} metric to define the ``difference'' between the real and synthetic domains by using the perception algorithm outputs. This metric is extended from the Contextualized Performance Difference (CPD) \cite{elmquist2022performance} and is similar to the metric used in \cite{hagn2021improved}. Unlike conventional approaches that focus on the performance of an algorithm trained on the synthetic dataset and tested on the real dataset, IPD focuses on performance similarity between instances tested on both synthetic and real datasets, making it a more accurate measure of simulation performance. The main idea of IPD is as follows: \textit{if a synthetic image closely resembles a real photo, when a perception algorithm poorly detects an object in the real photo, the same object in the synthetic image should also be detected poorly. Conversely, when the object is detected well in the real photo, the same should happen in the synthetic image}.

%% file: sections/related_works.tex
Learned Perceptual Image Patch Similarity (LPIPS), Peak Signal to Noise Ratio (PSNR), and Structural Similarity (SSIM) are typical metrics to measure performance of neural rendering. LPIPS measures the averaged response difference of the VGG16 layers between the matched input real and synthetic pictures \cite{zhang2018unreasonable}, and SSIM quantifies the similarity of two pictures based on their luminance, contrast, and composition \cite{wang2004image}. However, these two metrics focus on visual fidelity and image structural similarities itself, and these metrics require pixel-level matching between the real and synthetic pictures.

Metrics like Inception Score (IS) \cite{salimans2016improved}, Fréchet Inception Distance (FID) \cite{heusel2017gans}, and Kernel Inception Distance (KID) \cite{binkowski2018demystifying} are usually used to evaluate the generative adversarial network (GAN) performance. But they do not require exact matches between real and synthetic pictures, and they can not provide explicit evaluation for a specific perception algorithm. Other metrics, such as mean Average Precision (mAP), are often used to evaluate perception algorithms in object detection or semantic segmentation tasks \cite{tsirikoglou2017procedural, prakash2019structured}. However, using mAP to judge similarity between synthetic and real pictures can be misleading, as broader coverage of the synthetic domain could result in a higher mAP, even if the synthetic domain differs from the real domain.

%% file: sections/method.tex
\SetKwComment{Comment}{/* }{ */}
\begin{algorithm2e}[!ht]
	\caption{Calculation of Instance Performance Difference (IPD)}\label{alg:get_IPD}
	\KwIn{YOLOv5 as performance algorithm $H(image)$, labeled real image dataset $S_{real}$, labeled synthetic image dataset $S_{synth}$}
	Performance task: rock detection \\
	Performance-value: IOU \\
	$\bm{P}_{real} \gets$ empty\_list, $\bm{P}_{synth} \gets$ empty\_list\;
	\For{each pair of pictures $(I_{real}, I_{synth})$, $(I_{real} \in S_{real}, I_{synth} \in S_{synth})$}{
		\tcp{$n_{r}$ GT and $m_{r}$ predicted bboxes}
		predict\_bboxes $\bm{B}_{predict, real} \gets H(I_{real})$\;
		\tcp{$n_{s}$ GT and $m_{s}$ predicted bboxes}
		predict\_bboxes $\bm{B}_{predict, synth} \gets H(I_{synth})$\; 
		Get IOU table $T_{real} \in \mathbb{R}^{n_{r} \times m_{r}}$ by $\bm{B}_{predict, real}$\;
		Get IOU table $T_{synth} \in \mathbb{R}^{n_{s} \times m_{s}}$ by $\bm{B}_{predict, synth}$\;		
		\For{each pair of rocks $(x_{real}, x_{synth})$, $(x_{real} \in I_{real}, x_{synth} \in I_{synth})$}{
			$\bm{P}_{real}$ append $\max{\{T_{real}[x_{real},:]\}}$\;
			$\bm{P}_{synth}$ append $\max{\{T_{synth}[x_{synth},:]\}}$\;
		}
	}
	\KwRet{mean absolute error between $\bm{P}_{real}$ and $\bm{P}_{synth}$}
\end{algorithm2e}

\begin{figure*}[!t]
	\centering
	\includegraphics[width=1.6\columnwidth]{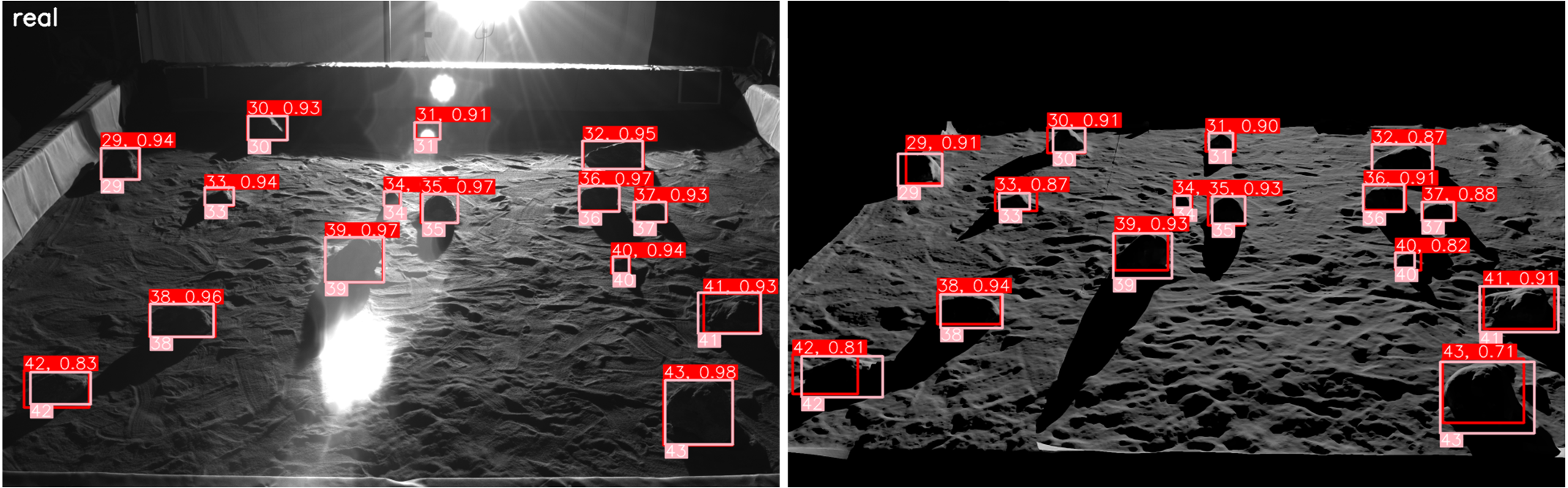}
	\caption{Indexed pairs of rocks from the (left) real and (right) synthetic pictures, respectively. The Pink boxes are the ground-truth labels with annotated rock indices, and the red boxes are the predicted labels with annotated rock indices and prediction confidence.}
	\label{fig:rock_index_pairs}
\end{figure*}

\begin{figure*}[!t]
	\centering
	\includegraphics[width=1.6\columnwidth]{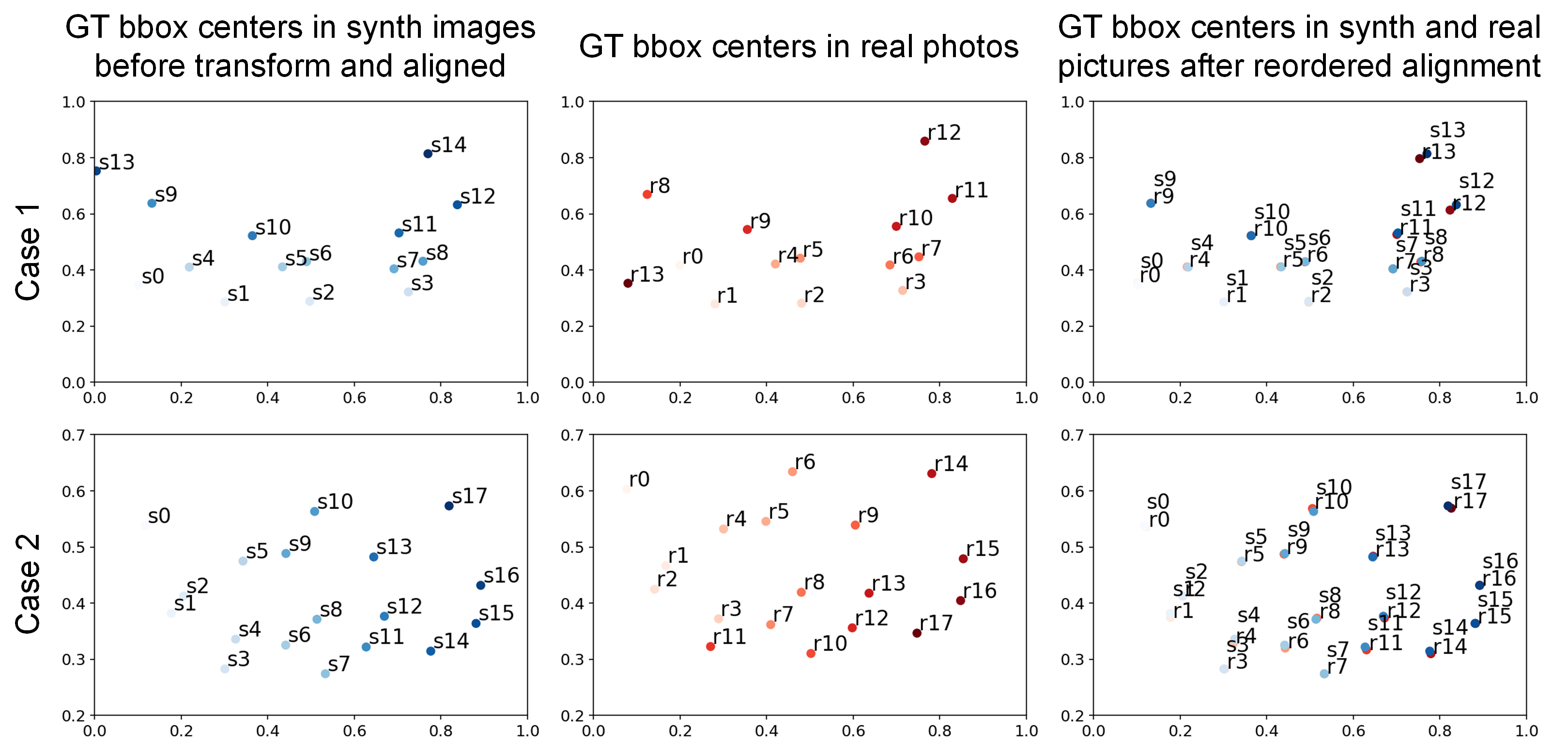}
	\caption{The point-set registration problem before and after solved of some cases shown for demonstration. (GT: ground-true, bbox: bounding box, synth: synthetic)}
	\label{fig:pt_reg_prblm}
\end{figure*}

In this technical contribution, we use the rock detection task for the synthetic images of the POLAR terrain scenarios \cite{wong2017polar} to explain the IPD computation method. The approach proposed to compute IPD is described in Algorithm~\ref{alg:get_IPD}, with YOLOv5 as the performance algorithm for evaluation in the rock detection. Intersection over Union (IOU) serves as the performance metric. Steps to calculate IPD between two datasets are as follows. First, synthetic and real rocks are paired and indexed, shown as the rock indices in Fig.~\ref{fig:rock_index_pairs}. Next, the trained YOLOv5 predicts bounding boxes for rocks in both real and synthetic testing sets. The predicted bounding box of the largest IOU with the ground-true (GT) label of the rock is chosen as the predicted label of the rock in either the real or synthetic set, as shown in Fig.~\ref{fig:rock_index_pairs}, and this IOU value is defined as the \textit{performance-value} of the rock. Lastly, the performance-value difference between the synthetic and real rocks in a pair is computed and averaged across all rock pairs. This averaged performance-value difference is defined as the IPD between the two datasets.

The ground truth (GT) bounding boxes of rocks are manually labeled in real photos and automatically generated in synthetic images. However, correctly pairing the corresponding synthetic and real rocks using these GT bounding boxes poses a problem. This is because some rocks are hidden by the wall's shadow in real photos but appear in the simulations, and the simulation setup does not perfectly align with reality. Thus, the number, order, and position of rocks in the pair of real and synthetic pictures are rarely aligned. This forms a 2D point-set registration problem, where the points are the GT bounding box centers. A modified RANSAC algorithm solves this problem. It iteratively finds the affine transformation matrix between all permutations of three randomly selected bounding box centers from the two respective datasets, and tries to achieve the minimal average Euclidean distance between the two aligned point sets after applying the transformation. The 2D point-set registration problem before and after solved is illustrated in Fig.~\ref{fig:pt_reg_prblm}.

%% file: sections/use_case.tex
A case is presented to demonstrate use of the IPD metric. In this case, we compare one real photo dataset and two synthetic image datasets related to the POLAR scenarios \cite{wong2017polar}: real photos from the POLAR dataset; images synthesized using the Principled Bidirectional Reflectance Distribution Function (BRDF) and assets in the POLAR-Sim dataset; and images synthesized using the Hapke BRDF \cite{hapke2008bidirectional, HapkeParams2014}. For more details, refer to the POLAR-Sim dataset paper \cite{chen2023polar3d}. We want to know which synthetic dataset is closer to the real dataset. We trained the YOLOv5 on each of the three datasets, respectively, and cross-validated on these three datasets by using the IPD metric. The results are shown in Table~\ref{table:IPD_result}.

\begin{table}[!t]
	\caption{Comparison of IPD results.}
	\label{table:IPD_result}
	\centering
	\resizebox{1.0\columnwidth}{!}{
		\begin{tabular}{|c|c|c|c|}
			\hline
			& \multicolumn{3}{c|}{$\downarrow$ Instance Performance Difference (IPD)} \\
			\hline
			Train\textbackslash Eval & $\|\text{Principled - Hapke}\|$ & $\|\text{Real - Hapke}\|$ & $\|\text{Real - Principled}\|$ \\
			\hline
			Real 		& - 				& 0.3152	& \textbf{0.2256}	\\
			\hline
			Principled	& \textbf{0.0511} 	& - 		& 0.3808			\\
			\hline
			Hapke		& \textbf{0.0261} 	& 0.4638	& - 				\\
			\hline
			\multicolumn{4}{l}{Note: the arrow pointing down $\downarrow$ in the first row means lower values are better}
		\end{tabular}
	}
\end{table}

\begin{figure}[!t]
	\centering
	\includegraphics[width=0.9\columnwidth]{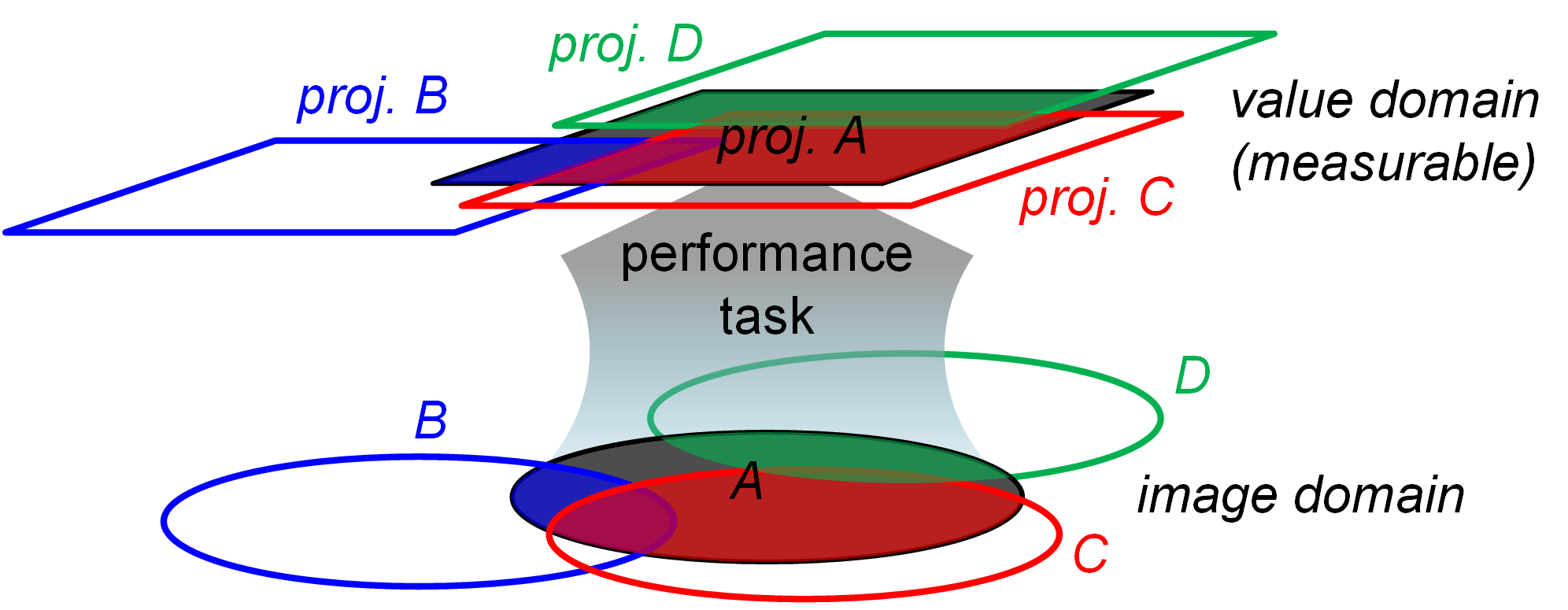}
	\caption{Illustration of IPD. The performance task trained on Domain A is treated as a mapping that projects from the image domains to the performance-value domains, which is only effective in Domain A.}
	\label{fig:IPD_illu}
\end{figure}

Note that Table~\ref{table:IPD_result}  ignores the IPDs that do not involve the training domain (i.e., the diagonal entries of the table). This is because, if we consider the performance task as a projection mapping from the image domain to the value domain -- with performance-values as observable data in the value domain -- it is meaningful only to consider differences within the data domain used for training this projection mapping. This approach is appropriate when comparing which testing data domain is closer to the training data domain, as illustrated in Fig.~\ref{fig:IPD_illu}.

Since the IPD, which was validated using the YOLOv5 trained on the real dataset (see the \textit{Real} row in Table~\ref{table:IPD_result}), between the real and Principled-synthetic datasets is smaller than the IPD between the real and Hapke-synthetic datasets, this result indicates that the Principled-synthetic images more closely resemble the real photos in the POLAR dataset compared to the Hapke-synthetic images.

%% file: sections/conclusions.tex
In this report, we proposed a new metric called \textit{Instance Performance Difference (IPD)}, a novel approach focused on perception algorithm performance to more accurately measure the sim-to-real gap between the synthetic and real picture datasets in robotic sensing simulations. This metric is particularly suited to the cases where synthetic images have one-to-one real counterparts and are used to train perception NNs. Unlike traditional metrics that primarily assess visual fidelity or statistical differences, IPD emphasizes performance similarity between the synthetic and real datasets, aiming to capture how effectively a synthetic image replicates real-world challenges in visual perception tasks. Through a use case leveraging the IPD metric with a rock detection task on the POLAR terrain dataset \cite{wong2017polar}, we demonstrated that the image dataset synthesized by the Principled BRDF aligns more closely with real-world data than the Hapke BRDF. This finding highlights the utility of the IPD metric in selecting and refining synthetic data generation techniques for training perception algorithms, ultimately supporting a more robust sim-to-real transfer in robotic applications.

Future work may further enhance the IPD metric to address more complex datasets and additional perception tasks, broadening its applicability across more diverse domains.